\documentclass[letterpaper]{article} 
\newcommand{\loss}{\mathcal{L}}
\newcommand{\E}{\mathbb{E}}
\newcommand{\R}{\mathbb{R}}

\newcommand{\N}{\mathcal{N}}
\newcommand{\X}{\mathcal{X}}

\newcommand{\pose}{\mathbf{x}}
\newcommand{\score}{\mathbf{s}_{\theta}}

\newcommand{\eps}{\epsilon}

\def\eg{\emph{e.g}.} 
\def\ie{\emph{i.e}.}

\usepackage{diagbox}
\usepackage{aaai24}  
\usepackage{times}  
\usepackage{helvet}  
\usepackage{courier}  
\usepackage[hyphens]{url}  
\usepackage{graphicx} 
\usepackage{xcolor}
\definecolor{green}{cmyk}{1,0.502,1, 0}
\definecolor{red}{RGB}{204,0,0}
\usepackage{natbib}  
\usepackage{caption} 
\usepackage{algorithmic}
\usepackage[ruled,vlined]{algorithm2e}
\usepackage{subcaption}

\usepackage[utf8]{inputenc} 
\usepackage[T1]{fontenc}    
\usepackage{booktabs}       
\usepackage{amsfonts}       
\usepackage{nicefrac}       
\usepackage{microtype}      
\usepackage{xcolor}         
\usepackage{graphicx}
\usepackage{multirow}
\usepackage{amsmath}        
\usepackage{mathrsfs}
\usepackage{array}

\usepackage{newfloat}
\usepackage{listings}
\DeclareCaptionStyle{ruled}{labelfont=normalfont,labelsep=colon,strut=off} 
\title{SocialGFs: Learning Social Gradient Fields for Adaptive Multi-Agent Systems}
\author {
    Qian Long\textsuperscript{\rm 1},
    Fangwei Zhong\textsuperscript{\rm 2},
    Mingdong Wu\textsuperscript{\rm 2},
    Yizhou Wang\textsuperscript{\rm 2},
    Song-Chun Zhu\textsuperscript{\rm 1, \rm 2}
}
\affiliations {
    \textsuperscript{\rm 1} University of California, Los Angeles\\
    \textsuperscript{\rm 2} Peking University\\
    longqian@ucla.edu, zfw@pku.edu.cn
}

\usepackage{bibentry}
\begin{document}

\maketitle

\begin{abstract}

Multi-agent systems (MAS) need to adaptively cope with dynamic environments, changing agent populations, and diverse tasks. However, most of the multi-agent systems cannot easily 
 handle them, due to the complexity of the state and task space. The social impact theory regards the complex influencing factors as forces acting on an agent, emanating from the environment, other agents, and the agent’s intrinsic motivation, referring to the social force. 
 Inspired by this concept, we propose a novel gradient-based state representation for multi-agent reinforcement learning.
 To non-trivially model the social forces, we further introduce a data-driven method, where we employ denoising score matching to learn the social gradient fields (SocialGFs) from offline samples, e.g., the attractive or repulsive outcomes of each force. During interactions, the agents take actions based on the multi-dimensional gradients to maximize their own rewards. In practice, we integrate SocialGFs into the widely used multi-agent reinforcement learning algorithms, e.g., MAPPO. The empirical results reveal that SocialGFs offer four advantages for multi-agent systems: 1) they can be learned without requiring online interaction, 2) they demonstrate transferability across diverse tasks, 3) they facilitate credit assignment in challenging reward settings, and 4) they are scalable with the increasing number of agents.

\end{abstract}

\section{Introduction}
A multi-agent system (MAS) is composed of multiple autonomous agents that interact with each other and the environment. MAS has many applications in domains such as games~\cite{vinyals2019grandmaster, openai2019dota}, social simulation~\cite{vinitsky2022learning, yaman2022emergence}, and distributed computing~\cite{shalevshwartz2016safe}. However, designing and learning MAS that can adapt to diverse and dynamic scenarios is a key challenge. For example, MAS may need to cope with changes in the environment (such as obstacles or resources) and the number, goals, and roles of agents (such as teammates or adversaries). Therefore, it is desirable to build a MAS that can transfer their knowledge across different settings. 

\begin{figure*}[t]
    \centering
    \vspace{-1mm}
        \includegraphics[width=\linewidth]{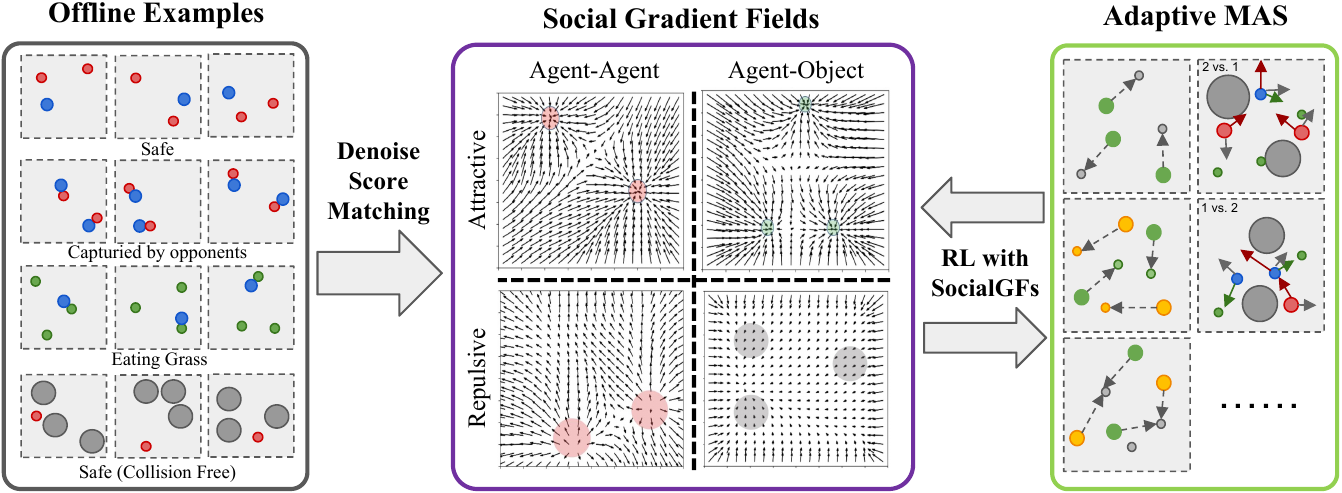}
        \vspace{-3mm}
        \caption{Learning Gradient Fields from Examples for Multi-agent Systems. We use a score matching function to train each example, obtaining different $gf$ functions. For various tasks, we select different sets of $gf$ and apply them to the observation to generate a gf-based representation (SocialGFs). We then apply RL methods to train the adaptive agent based on that representation. By employing different $gf$ functions for representation, the agent can adapt to various scenarios. }
        \label{fig:pipeline}
    \vspace{-0.15in}
\end{figure*}

Researchers on multi-agent reinforcement learning (MARL) have trained policies that can generalize across various aspects, such as multi-task performance~\cite{wen2022multi, omidshafiei2017deep}, scalability~\cite{agarwal2019learning, DBLP:journals/corr/abs-2003-10423, zhou2021cooperative, hu2021updet}, and communication \cite{omidshafiei2019learning, jiang2018learning, wang2022tomc}. Previous works have explored different techniques to enhance generalization, such as policy representation learning~\cite{grover2018learning}, meta learning~\cite{chen2021multiagent}, self-play~\cite{zhong2021towards}, curriculum learning~\cite{long2020evolutionary}, network architecture~\cite{jiang2018learning, xu2020learning}.
However, the agents learned by these methods are often tailored to specific tasks or environments and lack transferability and reusability. 

\begin{figure}[t]
    \centering
    \vspace{-1mm}
        \includegraphics[width=0.8\linewidth]{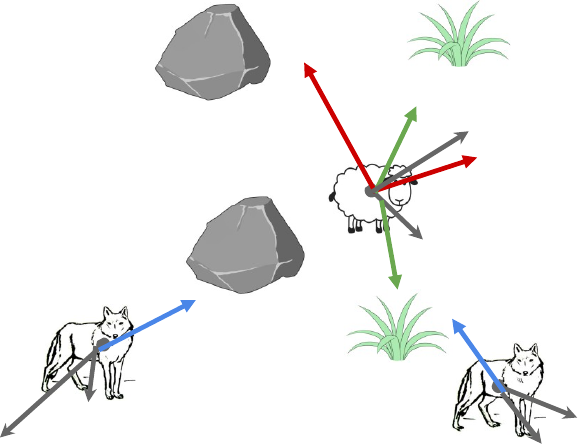}
        \vspace{-2mm}
        \caption{This is an example of the social forces that affect sheep and wolves in a grassland. The \textcolor{red}{red} arrows represent the forces from wolves that repel sheep, while the \textcolor{gray}{gray} arrows represent the forces from obstacles that prevent both sheep and wolves from escaping. The \textcolor{green}{green} arrow represents the force from the grass that attracts sheep, and the \textcolor{blue}{blue} arrows represent the forces from sheep that attract wolves. }
        \label{fig:force}
    \vspace{-0.2in}
\end{figure}

The most important part of achieving such adaption ability in multi-agent systems is to find a general and efficient way to represent complex environments.
The concept of social force~\cite{Helbing_1995}, borrowed from sociology, describes how individual behavior, interaction, and cognition are influenced by various factors in a social context. Examples of social force include attraction and repulsion among individuals, conformity and deviation from social norms, and cooperation and competition among groups. We argue that agents in a MAS are also subject to different types of forces that originate from the environment, other agents, and their intrinsic motivation. These forces can modulate the agents’ actions and strategies and ultimately determine their performance and adaptation. Fig.~\ref{fig:force} shows an example of the social forces of sheep and wolves in grassland.

The idea of using vector fields to represent forces has been explored in robotics, where artificial potential fields~\cite{warren1989global} or other gradient vector fields~\cite{zhao2022hybrid} have been applied to various tasks such as environment simulations~\cite{kolivand2021integration, wan2014research}, robot navigation~\cite{klanvcar2022robot, konolige2000gradient} and multi-agent path planning~\cite{matoui2019distributed, zheng2015multi}. However, these vector fields are often handcrafted and tailored to specific environments, limiting their generalizability and transferability. Furthermore, designing these vector fields explicitly is challenging in complex environments where multiple factors may influence the agents’ behavior. Therefore, it is necessary to \textbf{learn these vector fields from data rather than specify them manually}.

In this paper, we introduce Social Gradient Fields (SocialGFs), which are learned offline from examples of attractive or repulsive outcomes using denoising score matching~\cite{song2021scorebased}, a score-based generative modeling technique. SocialGFs represent these abstract forces as vector fields that guide the agents towards favorable or away from unfavorable states. 
Tasks can be represented as compositions of the gradients, and they often share common gradients, such as those for collision avoidance.
Hence, these gradients can be reused across different tasks. Moreover, as the gradient fields are dense, they can provide informative guidance in sparse reward scenarios, addressing the credit assignment problem in multi-agent learning.
Furthermore, with graph neural networks, they can also scale easily with the number of entities. With these features, we can unify the representation in multi-agent environments across varying tasks and populations. 

In the end, 

given these gradients to the agents, the agents only need to learn the policy from the gradient-based state representation in an RL manner, rather than learning a state representation from scratch. Since the gradient fields can provide generalist representation among tasks and scenarios, the agents can also adapt easily to a new environment by replacing the gradient-based representation. We show this pipeline in Fig.~\ref{fig:pipeline}.

To summarize, our main contributions are three-fold:
1) we propose learnable gradient fields, which are learned from offline examples by denoising score matching, for learning adaptive multi-agent policy.
2) we develop generalizable RL-based agents to learn to act based on the gradient fields and adapt to different scenarios.
3) we empirically demonstrate the effectiveness and generalization of our method in both the cooperative-competitive game and the cooperative game in a particle-world environment.

\section{Related work}

\textbf{Learning Adaptive Multi-agent System.}
Adaptive multi-agent systems consist of multiple agents that can learn from their interactions and adapt to changing environments and goals. One of the challenges is how to coordinate the agents to achieve a new goal while transferring to a novel environment. Agarwal et al.\cite{agarwal2019learning} used graph neural networks to handle the dynamic size of input. Long et al. ~\cite{DBLP:journals/corr/abs-2003-10423} used the attention mechanism and curriculum learning method to start from a small population environment and then adapt to large populations. Yang et al.\cite{yang2020mean} applied mean-field theory to incorporate with large number of entities. Zhou et al.\cite{zhou2021cooperative} and Hu et al.~\cite{hu2021updet} applied a transformer to deal with variant inputs. Liu et al.~\cite{liu2019value} used policy distillation \cite{rusu2016policy} for transfer learning among similar tasks. However, all of these methods rely on the high similarities between tasks. It would be hard for them to adapt to a completely new environment.

\textbf{Social Force in Multi-Agent System.} 

Social force model (SFM) is a powerful tool for simulating the behavior of agents in a variety of environments\cite{kolivand2021integration, wan2014research, chen2018social, huang2018social}. 
For example, Helbing et al.~\cite{Helbing_1995} used SFM to simulate pedestrian behavior in crowds, where pedestrians adjust their movement to reach a stable state in response to repulsive and driving forces. In this way, SFM replicates observed behaviors like lane formation and collision avoidance.
Gil et al.\cite{gil2021social} also applied SFM with machine learning techniques to build a social robot.  However, SFM is sensitive to the parameters of the models and is difficult to calibrate and validate. Another disadvantage of SFM is that they are not always generalizable to new environments. This is because SFM is usually manually designed and defined based on empirical observations of real-world agents, which makes it expensive or impossible to transfer. To address these limitations, we propose a new data-driven approach that is based on the examples and use the gradient field to represent the social force. This approach is more generalizable to complicated environments because it does not rely on empirical observations of real-world agents.

\textbf{Gradient Field for Decision Making} The Gradient field is a vector field that describes the trend of change of a function or a distribution. Thus, it can be used to represent the social force in a multi-agent environment. It has been widely applied for decision-making like navigation tasks~\cite{klanvcar2022robot, konolige2000gradient, guldner1995sliding}. Vail et al.~\cite{vail2003multi} applied gradient for role assignment based on different scenarios.  A recent work~\cite{zhao2022hybrid} introduced a hybrid gradient vector field for path-following. All of the above methods are planning-based methods that require knowing the explicit expression of the target function or distribution or being able to estimate or approximate it effectively. Another fold of methods is Learning-based methods. They learn a gradient field from data to guide decision-making, without requiring knowing the explicit expression of the target function or distribution. These methods usually use deep neural networks to fit an energy function or a score function, thus implicitly defining a gradient field. For example, TarGF~\cite{wu2022targf} uses denoising score matching~\cite{song2021scorebased} to learn target gradient fields to rearrange objects without explicit goal specification. It learns a score function by minimizing the squared difference between the score of an energy function and the score of the true data distribution. 
Building on this, GraspGF~\cite{wu2024learning}, GFPose~\cite{ci2022gfpose} and~\cite{xue2023learning} expand this methodology to broader applications such as dexterous hand manipulation, human pose estimation, and the irregular shape packing.
In this work, we also use denoising score matching to learn the gradient fields with a set of offline examples, but we use hybrid gradient fields to represent all the factors in the environments in multi-agent environments and further introduce an RL-based agent to act based on the gradient-based representation.

\section{Preliminary}

\subsection{Theory of Social Force}

Theory of social force~\cite{patten1896theory, Helbing_1995} assumes that agents are influenced by various social forces in their environment, such as attraction, repulsion, cohesion, or alignment. Social force theory can be formulated mathematically as a system of differential equations that describe the motion of each agent as a function of its position, velocity, and the forces exerted by other agents and external factors. The general form of the social force model is: 
\begin{equation}
    \frac{d\mathbf{v}_i}{dt} = f_i(\mathbf{x}_i,\mathbf{v}_i,\mathbf{x}_{-i},\mathbf{v}_{-i}), \frac{d\mathbf{x}_i}{dt} = \mathbf{v}_i
\end{equation}
where $\mathbf{x}_i$ and $\mathbf{v}_i$ are the position and velocity vectors of agent $i$, and $f_i$ is the net force acting on agent $i$, which depends on its own state and the state of other agents $(\mathbf{x}_{-i}, \mathbf{v}_{-i})$.
The specific form of $f_i$ can vary depending on the domain and the assumptions made about the agent's behavior and objectives. And the agents usually will be driven by multiple forces.

\subsection{Learning Gradient Fields via Score-Matching}
\label{sec:preliminary}
The score-based generative model aims to learn the \textit{gradient field} of a log-data-density, \ie, the \textit{score function}.
Given samples $\{ \pose_i \}_{i=1}^N$ from an unknown data distribution $\{ \pose_i \sim p_{\text{data}}(\pose) \}$, 
the goal is to learn a \textit{score function} to approximate $\nabla_{\pose} \log p_{\text{data}}(\pose)$ via a \textit{score network} $\score(\pose): \R^{|\X|} \rightarrow \R^{|\X|}$.
\begin{equation}
    \loss(\theta) = \frac{1}{2}\E_{p_{\text{data}}}\left[\left\Vert\score(\pose) - \nabla_{\pose} \log p_{\text{data}}(\pose)\right\Vert^2_2\right]
\label{eq: Vanilla Score-matching}
\end{equation}
During the test phase, a new sample is generated by Markov Chain Monte Carlo (MCMC) sampling, \eg, Langevin Dynamics (LD), which is out of our interest since we focus on gradient field estimation.

However, the vanilla objective of score-matching in Eq.~\ref{eq: Vanilla Score-matching} is intractable, since $p_{\text{data}}(\pose)$ is unknown. To this end, the Denoising Score-Matching (DSM)~\cite{denosingScoreMatching} proposes a tractable objective by pre-specifying a noise distribution $q_{\sigma}(\widetilde \pose|\pose)$, and train a score network to denoise the perturbed data samples, where $q_{\sigma}(\widetilde \pose|\pose) = \N(\widetilde \pose; \pose, \sigma^2I)$ is a Gaussian kernel with tractable gradient in our cases:
\begin{equation}
    \begin{split}
    \loss(\theta) &= \E_{\widetilde \pose \sim q_{\sigma}(\widetilde\pose|\pose), \atop \pose \sim p_{\text{data}}(\pose)}\left[\left\Vert\score(\widetilde \pose) - \nabla_{\widetilde \pose}\log q_{\sigma}(\widetilde \pose|\pose) \right\Vert^2_2\right] \\
    &= \E_{\widetilde \pose \sim q_{\sigma}(\widetilde\pose|\pose), \atop \pose \sim p_{\text{data}}(\pose)}\left[\left\Vert\score(\widetilde \pose) - \frac{1}{\sigma^2}(\pose - \widetilde \pose)
    \right\Vert^2_2\right]
\label{eq: DSM-Gaussian}
\end{split}
\end{equation}
DSM guarantees that the optimal score network holds $\score^*(\pose) = \nabla_{\pose} \log p_{\text{data}}(\pose)$ for almost all $\pose$.

In practice, we adopt an extension of DSM~\cite{song2020score} that estimates a \textit{time-dependent score network} $\score(\pose, t): \R^{|\X|} \times \R^{1} \rightarrow \R^{|\X|}$ to denoise the perturbed data from different noise levels simultaneously:
\begin{equation}
\begin{split}
     \loss(\theta) & = \E_{t\sim \mathcal{U}(\eps, T)}\\
    & \left\{\E_{
    \widetilde \pose \sim q_{\sigma(t)}(\widetilde\pose|\pose), 
    \atop
    \pose \sim p_{\text{data}}(\pose)}
    \lambda(t)\left[\left\Vert
    \score(\widetilde \pose, t) - 
    \frac{1}{\sigma^2(t)}(\pose - \widetilde \pose)
    \right\Vert^2_2\right] \right\}
\label{eq: SDE-Gaussian}
\end{split}
\end{equation}
where $T$, $\eps$, $\lambda(t) = \sigma^2(t)$, $\sigma(t) = \sigma_0^{t}$ and $\sigma_0$ are hyper-parameters. The optimal time-dependent score network holds $\score^*(\pose, t) = \nabla_{\pose} \log q_{\sigma(t)}(\pose)$ where $q_{\sigma(t)}(\pose)$ is the perturbed data distribution:
\begin{equation}
    q_{\sigma(t)}(\widetilde\pose) = \int q_{\sigma(t)}(\widetilde\pose|\pose) p_{\text{data}}(\pose) d\pose
\label{eq: perturbed_data_distribution}
\end{equation}

\vspace{-0.3cm}
{
\begin{algorithm}[tb]

 Generate $N_e$ examples from the environment $E(O_E,R_E)$ with the observation function $O_e$ and reward function $R_e$\;
 \While{$N<N_e$}{
  Learn gradient fields $gf$ from example $E_N$\;
  \eIf{$E_N\subseteq S^+$}{
   add $gf$ to attractive representation set $gf^+$\;
   \If{$R_E$ is sparse}{
 $R_E \gets R_E-\lambda \lvert gf^+\rvert$ \tcp*{credit assignment}
 }
 }
{add $gf$ to repulsive representation set $gf^-$\;
 }
 
 }
 $O_{GF}\gets\{gf^+,gf^-\}$\tcp*{combine GFs}
 \textbf{return} the new environment representation: $E(O_{GF},R_E)$\;

 \caption{Learning SocialGFs from Offline Examples}\label{alg:main}

\end{algorithm}
}

\section{Method}
In this section, we will introduce how to learn the gradient fields in the multi-agent environment and how to train an adaptive agent by using the gradient-based state representation.

\subsection{Problem Formulation}

In the context of a multi-agent environment denoted as $E(O_E, R_E)$, which is characterized by an observation function $O_E$ and a reward function $R_E$, we initially express $O_E$ as the concatenation of $gf$ vectors: $O_E$ is assigned as ${gf_1, gf_2, ..., gf_n}$. Additionally, we adapt the reward function $R_E$ to suit sparse reward environments by deducting the magnitude of the positive $gf^+$: $R_E$ is updated to $R_E - \lambda \lvert gf^+_E \rvert$.
For adaptation, we simply substitute the existing $gf$ representation, yielding the agent to respond accordingly to the new output. Further elaboration on these details follows.

 For representing the $O_E$ with $gf$. The initial step involves acquiring the example set ${S}$. For each example within this set, a score network $\score^*$ is trained by minimizing the expression specified in Equation \ref{eq: SDE-Gaussian}. This score network essentially embodies the learned gradient fields. By applying each $\score^*$ to the observation, a Gradient Field (GF) representation of the environment, denoted as $O_{GF_E}$, is obtained through the concatenation of all the individual outputs. This representation can be further subdivided into attractive GF $gf^+_E$ and repulsive GF $gf^-_E$ as $O_{GF_E}\gets\{gf^+_E,gf^-_E\}$. Thereby transforming the environment representation to $E(O_{GF_E}, R_E)$.

For scenarios with sparse rewards where exploration of the attractive example set is challenging, credit assignment is implemented. The magnitude of $gf^+_E$ is subtracted from these sparse attractive examples with a discount factor applied to the reward function $R_E$, resulting in $R_E \gets R_E - \lambda \lvert gf^+_E \rvert$.
Subsequently, Multi-Agent Reinforcement Learning (MARL) methods are applied to the new $gf$ representation of the environment. An policy trained on this represented environment $E$ is denoted as $\pi_\phi\{a_t|gf^+_E,gf^-_E\}$.

For adaptation, specifically transferring the policy $\pi_\phi\{a_t|gf^+_{E_1},gf^-_{E_1}\}$ trained on environment $E_1$ to the environment $E_2$, the GF representation $O_{GF_{E_2}}$ is obtained in the new environment, which is partitioned into attractive GF $gf^+_{E_2}$ and repulsive GF $gf^-_{E_2}$. Subsequently, the $gf^+_{E_1},gf^-_{E_1}$from $E_1$ are replaced with those from $E_2$, resulting in the updated agent $\pi_\phi\{a_t|gf^+_{E_2},gf^-_{E_2}\}$. $gf$ could be shared among different tasks.

\subsection{Offline Examples Collection}

The creation of example set ${S}$ involves the delineation of two fundamental categories: attractive examples ${S^+=\{s^+\}\subseteq S}$ and repulsive examples ${S^-=\{s^-\}\subseteq S}$ tailored to each type of agent based on the task description, reward function, and expert knowledge. Subsequently, these example sets will serve as the foundation for training gradient fields in subsequent stages.

To elucidate, in the guidance of agents towards task completion, attractive examples pertain to goal states, whereas repulsive examples encompass scenarios where agents get punished. Examples can be obtained from triggered event.
Let's consider a grassland game scenario, where the game dynamics stipulate that wolves prey on sheep, while sheep aim to evade wolves and consume grass. 
For the wolf, successful consumption of a sheep designates instances as attractive examples, whereas for the sheep, evading such situations constitutes repulsive examples. 
It is noteworthy that within each aspect, attractive ${S^+}$ or repulsive ${S^-}$, there may exist multiple types of example sets.
Different example sets correspond to distinct goal states that agents aim to achieve or avoid. This process of generating example sets proves particularly efficacious in scenarios with sparse rewards, where achieving the designated goal state poses a formidable challenge and the diversity of examples is inherently limited.

\begin{figure}[t]
\centering
        \includegraphics[width=\linewidth]{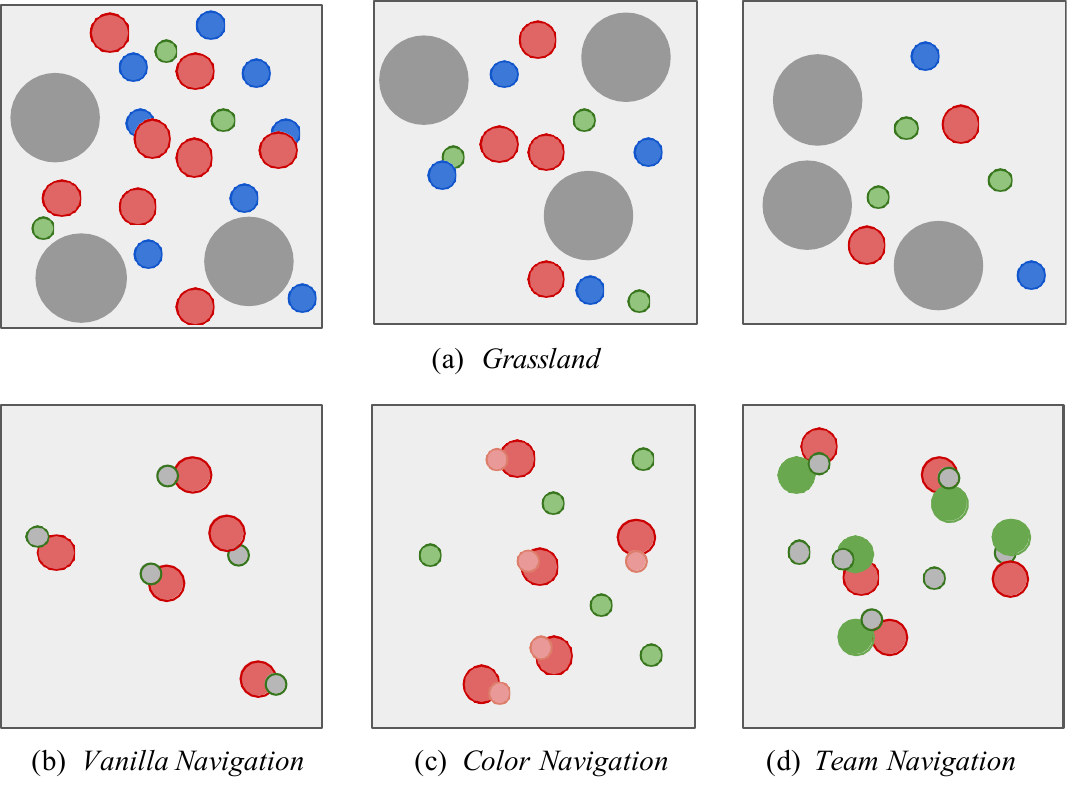}
        \vspace{-6mm}
        \caption{The offline examples that we used for learning gradient fields. The different colors of the balls indicate that they belong to different classes.}\label{fig:example}
\vspace{-0.3cm}
\end{figure}

In practice, we collect the examples based on the event triggered by the agents.
Fig.~\ref{fig:example} shows several examples extracted from the environments.
The different colors of the balls indicate that they belong to different classes.
We show examples of the grassland game in Fig.~\ref{fig:example}(a). The red ball represents a wolf and the gray ball represents sheep. The left example is collected when the ``sheep eaten" event is triggered. This is used as a repulsive example for sheep and a positive example for wolves. The middle example is collected when the ``grass eaten" event is triggered. This is used as a positive example for sheep. The right side example is collected from frames in the game. It is used as an agent-object example of legal position since the agent is not colliding with the boundaries or obstacles. It is used as a positive example for both sheep and wolves.
Fig.~\ref{fig:example} shows the three examples (b, c, d) of navigation games. They are collected when the ``success" event is triggered. 

\vspace{-0.2cm}
\subsection{Learning Gradient Fields}
Utilizing the attractive example sets ${S^+}$ and repulsive example sets ${S^-}$ as a foundation, we can proceed to train score networks to provide attractive gradient field $gf^+$ and repulsive gradient field $gf^-$, by adhering to the principles outlined in Equation \ref{eq: SDE-Gaussian}. The learned score networks can subsequently be employed at each step to estimate the gradient from each type of example set. This gradient serves as an elevated representation of the environment, offering insights into the direction and distance of an agent to all kinds of example sets within ${S}$.

\begin{figure}[t]
    \centering
        \includegraphics[width=0.9\linewidth]{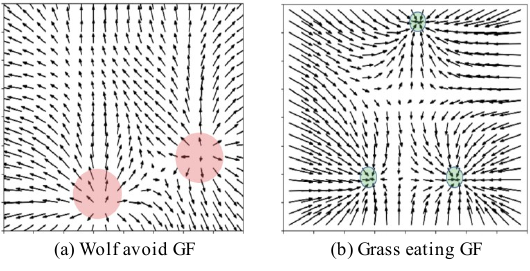}
        \vspace{-3mm}
        \caption{The visualization of learned Gradient fields for sheep in grassland game. Red and green circles indicate the wolves and the grass respectively.}
        \label{fig:gf}
    \vspace{-0.3cm}
\end{figure}

The gradient vectors obtained through this process encapsulate crucial information regarding an agent's proximity to attractive or repulsive instances, thereby enabling an abstract understanding of the environment. This higher-level representation, derived from the attractive and repulsive gradient field networks, enhances the agent's capacity to navigate and respond appropriately to the varying challenges posed by the attractive and repulsive examples within the environment.
In Fig.~\ref{fig:gf}, we show part of the $gf$ applied in the grassland game. The arrows are generated by putting sheep in that location. The length of the arrows measures the magnitude of the gradients. Fig.~\ref{fig:gf}(a) is the wolf avoid $gf_{wolf\_avoid}$ where the red circles are the positions of wolves. 
Fig.~\ref{fig:gf}(b) is the grass-eaten $gf_{grass\_eaten}$ where the green circles indicate grass. 

\subsection{Gradient-enhanced Rewards} In scenarios characterized by sparse rewards, where rewards are only allocated to a limited set of states, reward shaping becomes a common strategy. However, its applicability is problem-specific and demands meticulous engineering. In the context of $gf$ generation from these sparse attractive examples, the magnitude $\lvert gf^+\rvert$ serves as an indicator of the current state's distance from the distribution of attractive examples. Consequently, to mitigate sparsity, the absolute value of the GF, denoted as $\lvert gf^+ \rvert$, is subtracted from the reward function with a discount factor, resulting in $R_e \gets R_e - \lambda \lvert gf^+ \rvert$.

The objective is to maximize the reward function, thereby minimizing the distance between the current state and the example state. This approach ensures that the agent, by seeking to maximize rewards, progressively narrows the gap between its current state and the target state, facilitating more effective learning and convergence toward the desired outcomes.

\subsection{Learning Adaptive Policy}

Adaptive ability refers to an agent's capacity to alter its behavior in response to environmental shifts or novel tasks. The gradient field represents a form of representation adept at encoding the social rules and relationships within an environment. The introduction of adaptive ability in a Multi-Agent System can be accomplished through the replacement of $gf$ representation. An agent $\pi_\phi\{a_t|gf^+_{E_1},gf^-_{E_1}\}$ target for $E_1$ can transfer to $E_2$ by simply applying the $gf_{E_2}$ representation: $\pi_\phi\{a_t|gf^+_{E_2},gf^-_{E_2}\}$. A novel representation of the altered environment is introduced, compelling agents to transition to a new policy rooted in the updated $gf$. This transition prompts adjustments in the agents' interactions with each other and the environment, fostering a dynamic and responsive behavior reflective of the modified environmental conditions. Also, $gf$ can be reused among different tasks and agents.

\section{Experiment}
We tested our approach on four challenging environments, including a wolf-sheep game on grassland and three varieties of fully cooperative navigation games with sparse rewards. We found that our approach, SocialGFs, outperformed all other baselines in those four environments. We also tested the adaptive performance of SocialGFs across tasks and with different scales of agent populations and found that it performed well in those four environments.

\subsection{Environments}

All of the environments we used in our experiments were built on top of the particle-world environment ~\cite{mordatch2017}, which is a continuous 2D world where agents can take actions in discrete timesteps.

\begin{figure}[t]
    \centering
        \includegraphics[width=\linewidth]{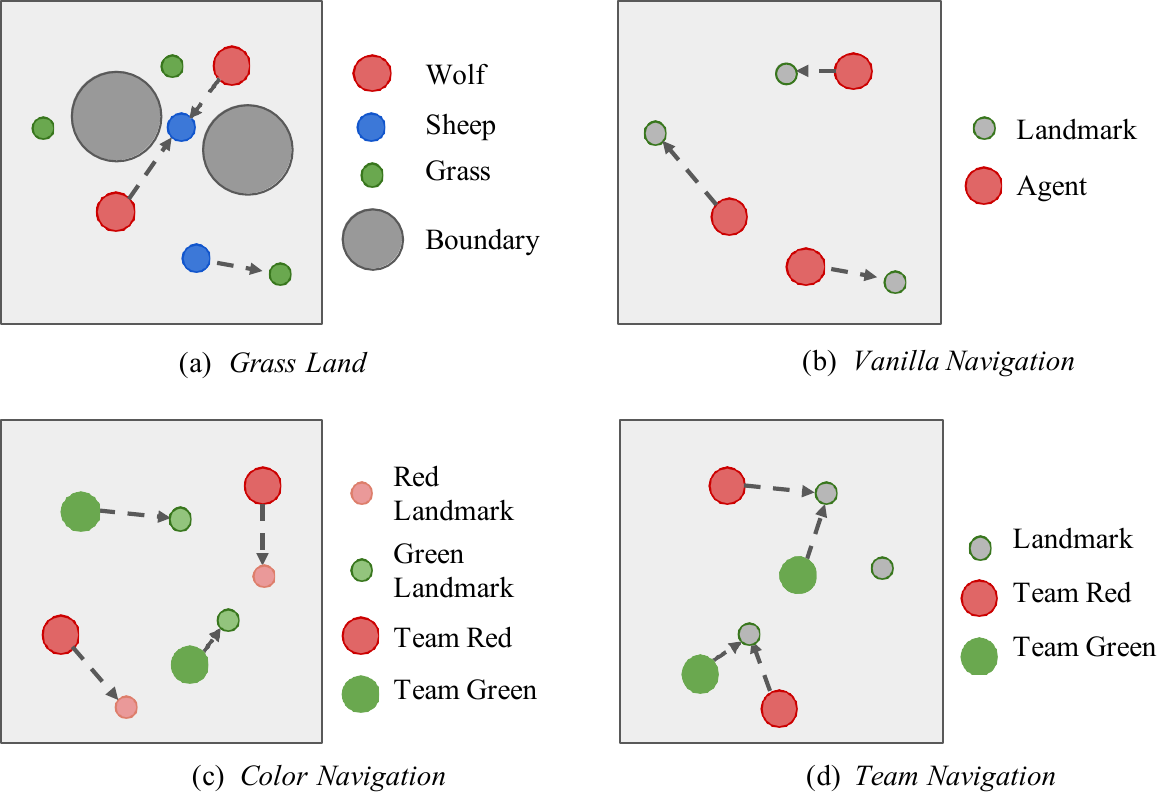}
        \vspace{-4mm}
        \caption{Four games used in the experiments. Figure (a) shows the grassland game where sheep work together to collect grass and avoid wolves, while the wolves work cooperatively to eat sheep. Figure (b)(c)(d) are three varieties of cooperative navigation games where agents need to cooperatively reach different landmarks.}\label{fig:env}
\end{figure}

\noindent\textbf{Grassland:} 
 As is shown in Fig.~\ref{fig:env}(a), there are two types of agents: sheep (blue) and wolves (red). The sheep need to collect grass pellets (green) and avoid wolves, while the wolves need to eat sheep. The sheep move faster than the wolves. There are a fixed number of grass pellets (food for sheep) and large gray obstacles in the environment that can impede the movement of the agents.
When a wolf collides with (eats) a sheep, the wolf is rewarded, and that (eaten) sheep will obtain a penalty. This will be marked as a ``sheep eaten" event. When any sheep comes across a grass pellet, that sheep is rewarded and the grass will be collected and respawned in another random position, and this will be marked as a ``grass eaten" event.
The goal of the sheep is to collect as many grass pellets as possible and avoid being eaten by the wolves. The goal of the wolves is to eat as many sheep as possible.

\noindent\textbf{Cooperative Navigation:}
In this environment, agents must cooperate through physical actions to reach a set of landmarks $L$. The agents will receive a reward and the event will be marked as ``success" only when all the landmarks are occupied correctly. We designed three varieties of cooperative navigation for different difficulties:
\begin{itemize}

    \item \textbf{\emph{Vanilla Navigation}}: $N$ cooperative agents aim to simultaneously occupy $N$ landmarks without any conflicts, as is shown in Fig.~\ref{fig:env}(b).
    \item \textbf{\emph{Color Navigation}}:
    There is an equal number of red and green agents, as well as an equal number of corresponding red and green landmarks. The objective of the game is for each agent to navigate towards and reach the landmark that matches their own color, as is shown in Fig.~\ref{fig:env}(c). 
    \item \textbf{\emph{Team Navigation}}: In the game shown in Fig.~\ref{fig:env}(d), there are two teams of agents, one red and the other green. Both teams are composed of an equal number of agents. However, the number of agents exceeds the number of available landmarks. For a landmark to be deemed successfully occupied, it must be touched by at least one agent from each team.
\end{itemize}
\vspace{-0.15in}

\subsection{Baselines and Evaluation Method}
 In the experiments, we compared the following approaches: 
 \begin{itemize}
     \item \textbf{Original(sparse) Reward}: this method trains the agents using event-based sparse reward and MAPPO ~\cite{yu2021surprising} algorithm; 
     \item \textbf{Reward Engineering}: this method trains the agents with human-shaped dense reward and MAPPO; 
     \item \textbf{SocialGFs(*)}: Agents are trained with the original reward using the gradient fields learned from previously collected offline examples. The adaptive model, denoted by an asterisk (*), is trained from the SocialGFs sheep model at a scale of $4-4$ within the Grassland game environment.
     \item \textbf{SocialGFs+} (for Cooperative Navigation): Specifically designed for sparse reward scenarios, this method further incorporates gradient field representation with credit assignment.
 \end{itemize}

For the grassland, we follow the evaluation method in \cite{DBLP:journals/corr/abs-2003-10423} by competing agents trained from different methods. Specifically, we let sheep trained by different approaches compete with wolves trained from SocialGFs to get the average reward as the evaluation of sheep. Similarly, we compete for wolves with SocialGFs sheep to evaluate different wolves. For better visualization, we use normalized reward from 0.1-1 to show the difference between methods. We also do a cross-match of every wolf and sheep trained under a scale of 4-4. In cooperative navigation, we calculate the final step success rate as evaluation metrics.

\subsection{Quantitative Results in \emph{Grassland}}

We denote a game with $N_W$ wolves and $N_S$ sheep by ``scale $N_W$-$N_S$''. Both $N_W$ and $N_S$ can choose from $\{2,4,8\}$ so totally there are 9 scales. 
The SocialGFs sheep's GF-representation is: $\{gf^+_{grass\_eaten}, gf^+_{boundary\_avoid}, gf^-_{wolf\_avoid}\}$, and the wolf's GF-representation is: $\{gf^+_{sheep\_chasing}, gf^+_{boundary\_avoid}\}$. During MARL training, every method is trained with $2\times 10^6$ episodes on all scales. The SocialGFs* we choose are sheep models trained from scale $4-4$. To adapt the sheep, there is nothing to change since the $gf$ representation is the same. For wolves, we replace $gf^+_{grass\_eaten}$ with the wolves $gf^+_{sheep\_chasing}$ and remove the $gf^+_{grass\_eaten}$.

\begin{figure}[!tb]
    \centering
        \includegraphics[width=0.99\linewidth]{ 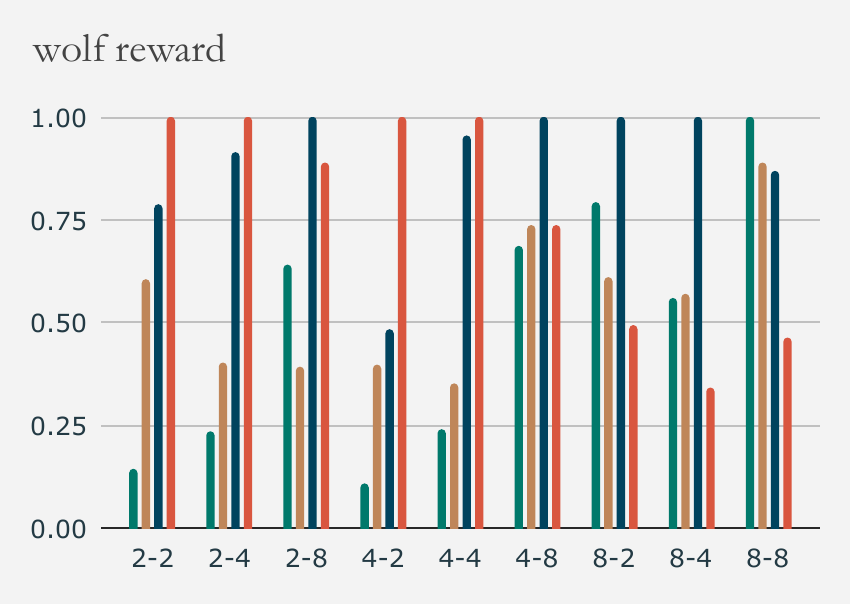}
        \includegraphics[width=0.99\linewidth]{ 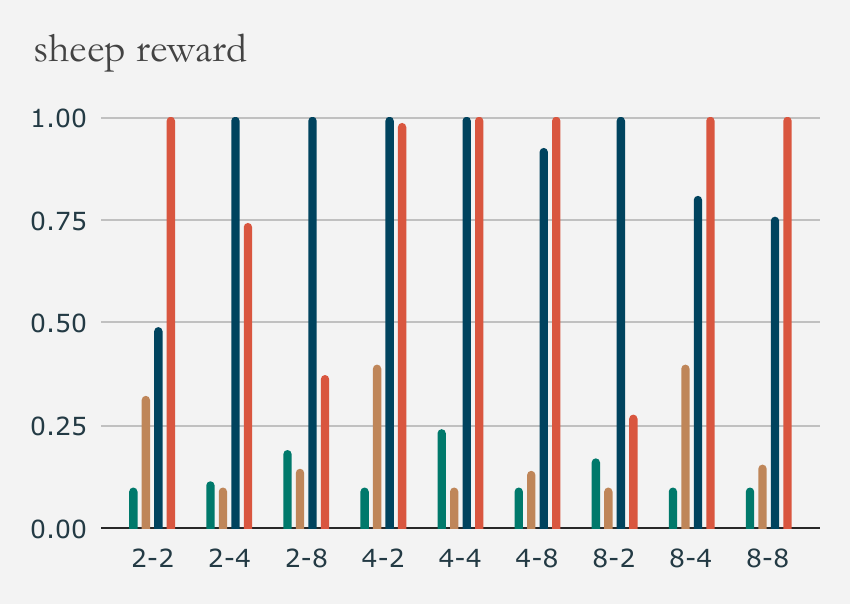}
        \includegraphics[width=0.99\linewidth]{ 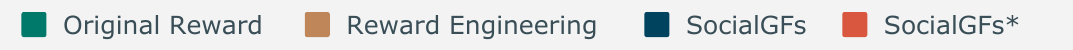}
        \caption{The quantitative results in Grassland. In the left two figures, we scale the wolf and sheep reward to 0.1-1. In the rightmost figure, we list the grass eaten amount for sheep in 100 time steps to measure the grass collection ability.}
        \label{fig:grasslandscore}
        \vspace{-0.3cm}
\end{figure}

\noindent\textbf{Main Results:}
The results are shown in Fig.\ref{fig:grasslandscore}. From the graph, we can see that SocialGFs* and SocialGFs dominate most of the scales. Overall, there is little difference between the original reward and Reward Engineering results. For adaptive ability, SocialGFs* adapt perfectly as sheep for all the scales, surprisingly, they are even the best wolves for a few scales. We also do cross matches between all the wolves and sheep trained from different methods on scale of 4-4. The result is shown in Table~\ref{table:grassland}. The red numbers are the reward of wolves and the green numbers are the result of sheep. When we fix the same group of sheep or wolves, the SocialGFs and SocialGFs* method always perform better compared to the other methods.

\begin{table}[t]
\centering
\caption{The cross-validation results (rewards) on \emph{Grassland} with 4 sheep and 4 wolves.}
\resizebox{0.49\textwidth}{!}{
\begin{tabular}{|c|c|c|c|c|}\hline
\backslashbox{\textcolor{green}{Sheep}}{\textcolor{red}{Wolf}}
&Original Reward&Reward Engineering&SocialGFs&$SocialGFs^*$\\\hline
Original Reward &\backslashbox{\textcolor{green}{-0.62}}{\textcolor{red}{0.904}}&\backslashbox{\textcolor{green}{-0.638}}{\textcolor{red}{0.905}}&\backslashbox{\textcolor{green}{-1.819}}{\textcolor{red}{2.351}}&\backslashbox{\textcolor{green}{-0.734}}{\textcolor{red}{1.07}}\\\hline
Reward Engineering &\backslashbox{\textcolor{green}{-1}}{\textcolor{red}{1.345}}&\backslashbox{\textcolor{green}{-0.316}}{\textcolor{red}{0.662}}&\backslashbox{\textcolor{green}{-3.86}}{\textcolor{red}{4.804}}&\backslashbox{\textcolor{green}{-1.894}}{\textcolor{red}{2.577}}\\\hline
SocialGFs &\backslashbox{\textcolor{green}{0.323}}{\textcolor{red}{0.169}}&\backslashbox{\textcolor{green}{0.209}}{\textcolor{red}{0.266}}&\backslashbox{\textcolor{green}{0}}{\textcolor{red}{0.695}}&\backslashbox{\textcolor{green}{-0.178}}{\textcolor{red}{0.855}}\\\hline
$SocialGFs^*$&\backslashbox{\textcolor{green}{0.315}}{\textcolor{red}{0.169}}&\backslashbox{\textcolor{green}{0.206}}{\textcolor{red}{0.273}}&\backslashbox{\textcolor{green}{-0.019}}{\textcolor{red}{0.707}}&\backslashbox{\textcolor{green}{-0.192}}{\textcolor{red}{0.877}}\\\hline
\end{tabular}
}
\label{table:grassland}
\end{table}

\begin{table*}[t]
\centering
\caption{The success rate on three cooperative navigation games with different populations.}
\resizebox{\textwidth}{!}{%
\begin{tabular}{|c|c|c|c|c|c|c|c|c|c|c|c|c|}\hline
   {}&\multicolumn{4}{|c|}{Vanilla Navigation}&\multicolumn{4}{|c|}{Color Navigation}&\multicolumn{4}{|c|}{Team Navigation}\\\hline
   {$\#$ of agents}&2&3&4&5&2&3&4&5&2&3&4&5\\\hline
      Original Reward& 0.01&0.0&{0.0}&0.0 & 0.0&0.0&{0.0}&0.0 & 0.0&0.0&{0.0}&0.0\\\hline
      Reward Engineering & 0.19&0.0&{0.0}&0.0 & 0.05&0.0&{0.0}&0.0 & 0.004&0.001&{0.0}&0.0 \\\hline
      SocialGFs & \textbf{0.998}&0.948&{0.88}&0.0 & 0.0&0.0&{0.0}&0.0 & 0.0&0.0&{0.0}&0.0 \\\hline
      $SocialGFs^+$ & {0.992}&\textbf{0.971}&\textbf{0.883}&\textbf{0.751} & \textbf{0.484}&\textbf{0.428}&\textbf{0.348}&\textbf{0.240} & \textbf{0.359}&\textbf{0.265}&\textbf{0.203}&\textbf{0.106} \\\hline
      $SocialGFs^*$ & 0.771&0.639&{0.571}&0.400 & 0.417&0.302&{0.243}&0.142 & 0.121&0.098&{0.051}&0.038 \\\hline
\end{tabular}%
}
\vspace{-0.3cm}
\label{table:navigation}
\end{table*}

\subsection{Quantitative Results in \emph{Cooperative Navigation}}

In the vanilla navigation game, we have 2-5 agents and the number of landmarks is the same as the agent. For the color navigation game, there are two teams of the same equal agents, each with 2-5 agents and the landmark number is equal to the number of agents. For the team navigation game, there are two teams of equal agents. The number of agents in each team and landmarks are (2,3), (3,5), (4,6), and (5,7).
The SocialGFs agent's GF-representation is $gf^+_{navigation}$. The SocialGFs* we choose are sheep models trained in a grassland game of scale 4-4. We replaced the representation from $gf^+_{grass\_eaten}$, $gf^+_{boundary\_avoid}$, $gf^-_{wolf\_avoid}$ with $gf^+_{navigation}$.

We list the success rate of different navigation games in Table~\ref{table:navigation}. Both the original reward method and the Reward Engineering method failed in all scenarios except the smallest scale of the vanilla navigation game. SocialGFs are only able to work in simpler scenarios in vanilla navigation games that have less than 5 agents. SocialGFs+ outperforms all the other methods with a large gap in all the environments. SocialGF* agents are also managed to succeed in all the cooperative navigation games which shows the strong adaptability of SocialGFs.

\begin{figure}[!t]
\centering
        \includegraphics[width=\linewidth]{ 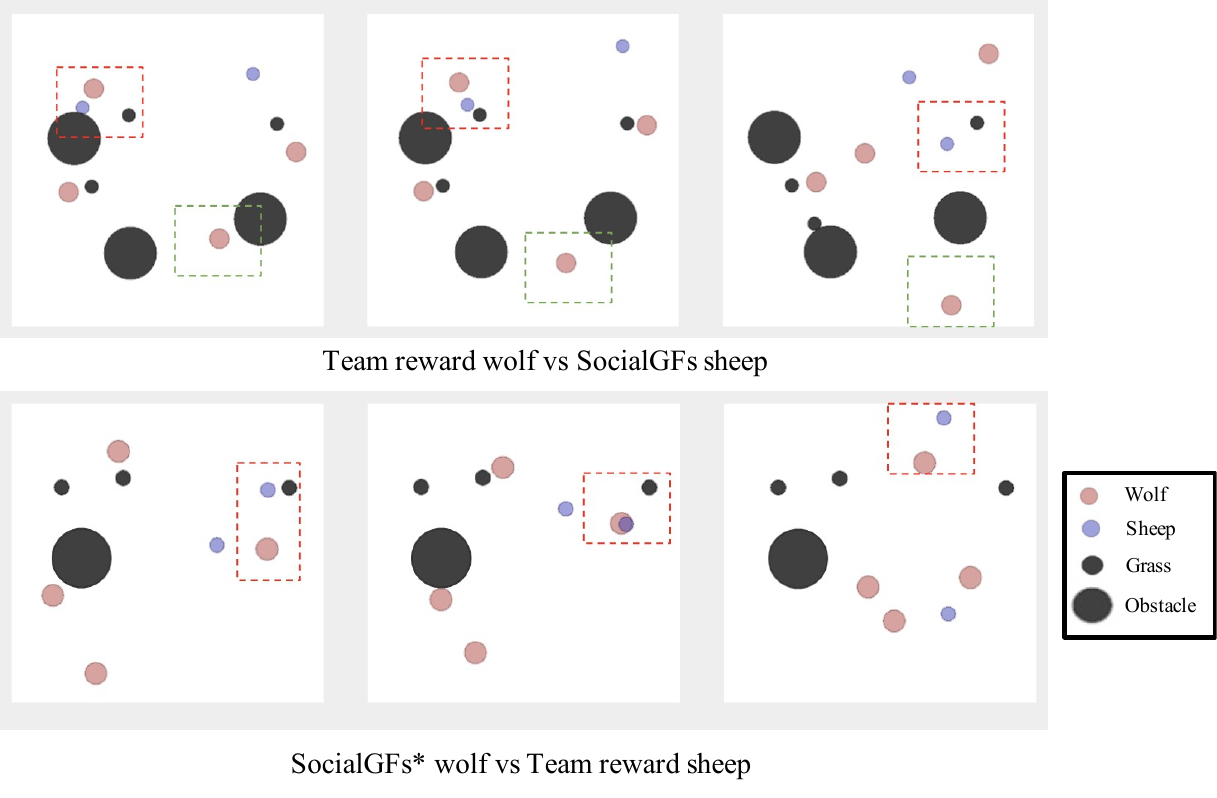}
        \vspace{-5mm}
        \caption{The qualitative results in grassland game. We visualize two representative grassland matches between SocialGFs agents and original reward agents. The upper graph shows the results of a simulation with 4 original reward wolves and 2 SocialGF sheep. The green box shows that one of the wolves is not chasing any sheep and the other wolves are not able to catch the sheep. The red box shows a sheep can collect grass and avoid the wolves at the same time.
The lower graph shows the results of a simulation with 4 SocialGFs* wolves and 2 original reward sheep. The red box shows that the wolves can successfully corner and eat the sheep.}\label{fig:glq}

    \vspace{+3mm}
        \includegraphics[width=\linewidth]{ 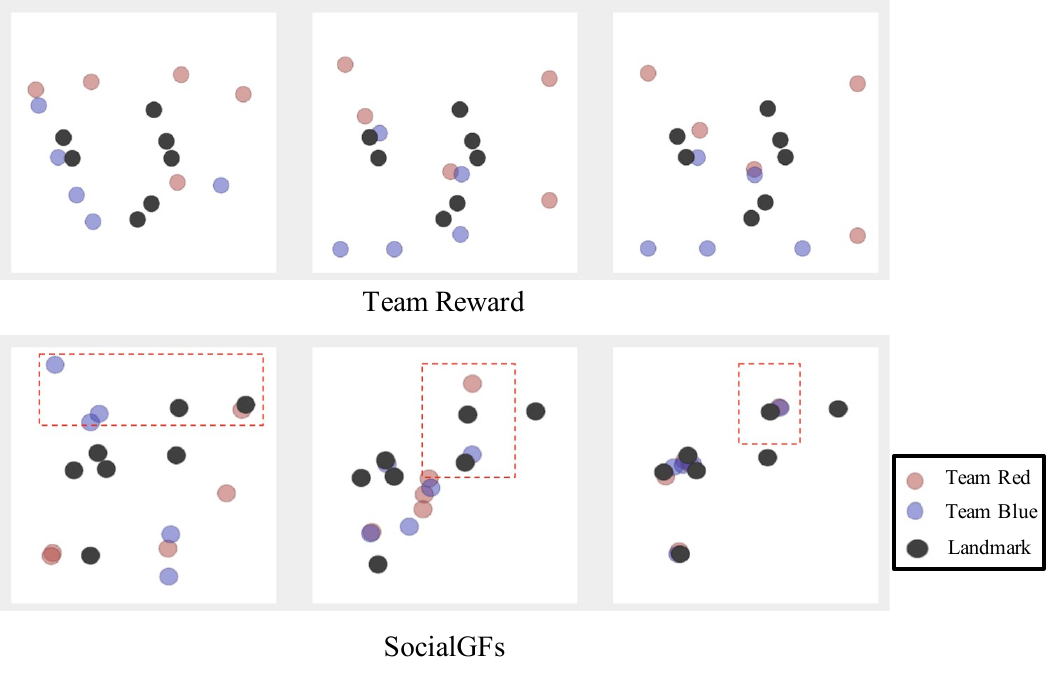}
       \vspace{-5mm}
        \caption{The qualitative results in team navigation game. We visualize two representative team navigation
games from the original reward method and SocialGFs method. The upper graph shows that the original reward method is unable to learn to play this game. The bottom figure shows the SocialGFs agents successfully cooperate together to occupy 5 landmarks}\label{fig:cnq}
\end{figure} 

\subsection{Qualitative Results}

In the grassland game, as the number of wolves increases, it becomes more difficult for sheep to develop grass-collecting abilities. This is because the wolves are able to more easily catch and eat the sheep, which reduces the amount of time that the sheep have to collect grass.

We take a few example screenshots of two representative grassland matches between SocialGFs agents and original reward agents from the experiment rendering result in Fig.~\ref{fig:glq}. The small gray balls represent grass. The red balls represent wolf and the blue dot represents sheep.   The upper graph of Fig.~\ref{fig:glq} shows the results of a simulation with 4 original reward wolves and 2 SocialGFs sheep. The green box shows that one of the wolves is not chasing any
sheep and the other wolves are not able to catch the sheep. The red box shows a sheep can collect grass and avoid the wolves at the same time. The lower graph shows the results of a simulation with 4 SocialGFs* wolves and 2 original reward sheep. The red box shows that the wolves can successfully corner and eat the sheep.

We also visualize two representative team navigation games in Fig.~\ref{fig:cnq}. Among the cooperative navigation games, the team navigation game is the most difficult. The upper graph of Fig.\ref{fig:cnq} shows that the original reward method is unable to learn to play this game. The bottom figure shows the SocialGFs agents successfully cooperating to occupy 5 landmarks. In the red box, we show two agents from different teams coming all the way together to occupy the landmark in between.

These results suggest that agents trained with SocialGFs have adapted successfully in new environments and SocialGFs is a more effective method for training agents to play the grassland game and cooperative navigation games than the original reward method.

\section{Conclusion}
In this paper, we proposed the learnable gradient-based state representation, namely \textit{SocialGFs}, for building adaptive multi-agent systems. SocialGFs are learned from a set of collected exemplar states by denoising score matching. Using the GFs as the new representation to enhance the state and reward, the agents can efficiently learn the policy via MARL. Our approach has demonstrated significant improvement over the baseline. We also show the powerful adaptive ability of the SocialGFs-based agents across tasks. Given these encouraging results, we believe that our work has provided a new perspective to enhance the adaptive ability of multi-agent systems.

\textbf{Limitations} and \textbf{Future works}: 
In future work, we can rank the importance of each GF at the start of generalization to a new environment for better adaptation across tasks. Extending the SocialGFs to more photo-realistic 3D environments, e.g., UnrealCV~\cite{qiu2017unrealcv}, is also one of our future work.

\section{Board Impact}
The broad impact of this work is that it provides a new and promising approach to MARL. SocialGFs can get a more abstract and scalable representation of environments than traditional MARL methods, and they can be used to train agents that can adapt to changing environments, agent populations, and state spaces. This work has the potential to make a significant impact on the development of autonomous systems that can interact with other agents in complex and dynamic environments.
Here are some specific examples of how SocialGFs could be used to improve the performance of MARL systems in real-world applications:

\paragraph{Self-driving cars:} SocialGFs could be used to train self-driving cars to interact safely with other vehicles and pedestrians. SocialGFs could be used to model the forces of other vehicles or pedestrians when they are crossing the street. By applying the repulsive GFs the SocialGFs trained agent can act more effectively under dangerous scenarios.

\paragraph{Robotics:} SocialGFs could be used to train robots to interact safely and effectively with humans. That involves a lot of different tasks and goals that robots need to adapt to. Thus, by using SocialGFs, we can build an adaptive robot that can work around in large variety of scenarios.

\bibliography{aaai24}

\newpage
\appendix
\onecolumn
\section{Algorithm Details}
\begin{algorithm}

 Given a set of example $S$;\\
\For {example $e$ in $S$}{
   Categorize $e$ to example set $E_i$ based on triggered events $e_i$\tcp*{Categorize each example}}
Train different categories of target score network $\Phi_i$ via Denoise Score Matching (Alg.\ref{alg:main}) on all the example set $E_i$;\\
 Get $gf$ representation of the environment: $O_{GF}\gets\{gf^+,gf^-\}$ ;\\
 Conduct the MARL training on the $gf$ represented environment: $E(O_{GF},R_E)$\;

 \caption{Whole Training Process}\label{alg:complete}

\end{algorithm}
Algorithm 2 conducts the entire training process. It begins by categorizing each example from the given set S into specific example sets $E_i$ based on triggered events $e_i$. Subsequently, it trains various categories of target score network $\Phi_i$ using Denoise Score Matching (Alg.\ref{alg:main}) on all the example sets $E_i$. Next, it obtains a representation of the environment, denoted as $O_{GF}$, which consists of positive ($gf^+$) and negative ($gf^-$) gradients. Finally, it carries out MARL training on the environment represented by $O_{GF}$ and the reward function $R_E$.
\section{Environment Details}
\subsection{Reward}
In the grassland game, the sheep gets $+2$ reward when the sheep eat the grass, $-5$ reward when eaten by the wolf. The wolf gets $+5$ reward when eats a sheep. For the \emph{Reward Engineering}, the wolf's reward will be minus the minimum distance to the sheep, $R_{wolf}-=Min(Distance(self, sheep_{all}))$ and the sheep's reward will be minus the minimum distance to the grass: $R_{sheep}-=Min(Distance(self, grass_{all}))$.

In the cooperation navigation games, every agent will get a reward of 10 after every landmark is correctly occupied. For the reward engineering method: the agent reward will be minus the minimum distance from each landmark to that agent: $Reward -= Min(Distance(self, landmark_{all}))$. The agent will also get a bonus +1 reward when it successfully occupies a landmark.

\subsection{Observation}

In the grassland game, agents receive comprehensive information including the relative positions of other agents, grass, and obstacles, alongside the velocities of the other agents. In cooperative navigation games, in addition to the relative locations and velocities of other agents, they also gain access to color markings and relative locations of all other landmarks.

\subsection{Action}

The agent's action is represented by a two-dimensional continuous vector, which describes the force applied to an entity, considering both magnitude and angular direction.

\section{Training Details}
We follow all the hyperparameters in \cite{wu2022targf} for GF learning based on Table~\ref{table:targf} and PPO~\cite{yu2021surprising} for MARL (Table~\ref{table:mappo}). We select 1000 examples for learning every $gf$ and set $t=0.01$ for computing the $gf$ score. 

\begin{center}
\begin{table}[ht]

\caption{The hyperparameters for learning GF. }
\centering
\begin{tabular}{||c c c c c c c c||} 
 \hline
 $lr$ & activation & sigma & $t_0$ & hidden size & optimizer & optimizer betas & network\\  
 \hline\hline
 $2e-4$ & Relu & 25 & 1 & 64 & Adam & $[0.5,0.999]$ & GNN\\ 
 \hline
\end{tabular}
\label{table:targf}
\end{table}
\end{center}

\vspace{-12mm}
\begin{center}
\begin{table}[ht]
\centering
\caption{The hyperparameters for MAPPO.}
\resizebox{\textwidth}{!}{%
\begin{tabular}{||c c c c c c c c||} 
 \hline
 $lr$ & activation & gain & share policy & hidden size & optimizer & optimizer epsilon & network\\  
 \hline\hline
 $7e-4$ & Tanh & 0.01 & False & 64 & Adam & $1e-5$ & MLP\\ 
 \hline
\end{tabular}%
}

\label{table:mappo}
\end{table}
\end{center}

\section{Network Structure}
\begin{figure}[ht]

\centering
        \includegraphics[width=1\linewidth]{ 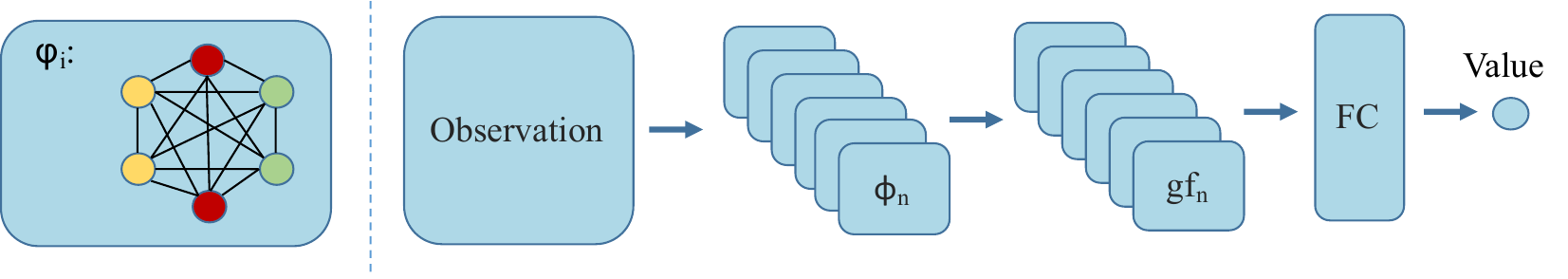}
        \caption{The structure of the Neural Network. We delve into the network architecture devised for crafting $gf$ representations and driving the generation of the final action. The target score network functions $\phi_i$ harness the capabilities of graph neural networks to adeptly capture intricate relationships among the various entities, where nodes symbolize both agents and landmarks. We construct the gradient field representation $O_{GF}$ by applying each $GF_i$ function $\phi_i$ to its corresponding category i and consolidating the resulting outputs $gf_i$ through concatenation. This concatenated representation $O_{GF}$ undergoes further refinement through a fully connected layer FC, culminating in the production of the final action. }
        \label{fig:structure}

\end{figure}

In Section 9, we delve into the network architecture devised for crafting $gf$ representations and driving the generation of the final action. Specifically, the target score network functions $\phi_i$ harness the capabilities of graph neural networks to adeptly capture intricate relationships among the various entities, where nodes symbolize both agents and landmarks. 
The target score network seamlessly embeds observations into distinct gradient fields $gf_i$. The structure of $\phi$ is based on the codes from the TarGF \cite{wu2022targf}. It contains two hidden dimensions with a hidden dimension size of 64 and then connects to the EdgeConv layer to get the final output.

Throughout the process of MARL training, we construct the gradient field representation $O_{GF}$ by applying each $GF_i$ function $\phi_i$ to its corresponding category i and consolidating the resulting outputs $gf_i$ through concatenation. Subsequently, this concatenated representation $O_{GF}$ undergoes further refinement through a fully connected layer (FC), culminating in the production of the final action. FC is identical for all the tasks based on the codes from the MAPPO \cite{yu2021surprising}. It has two hidden layers and the hidden dimension is 64.

\section{Evaluation Results Details}

\begin{figure}[ht]
    \centering
    \vspace{-1mm}
        \includegraphics[width=0.5\linewidth]{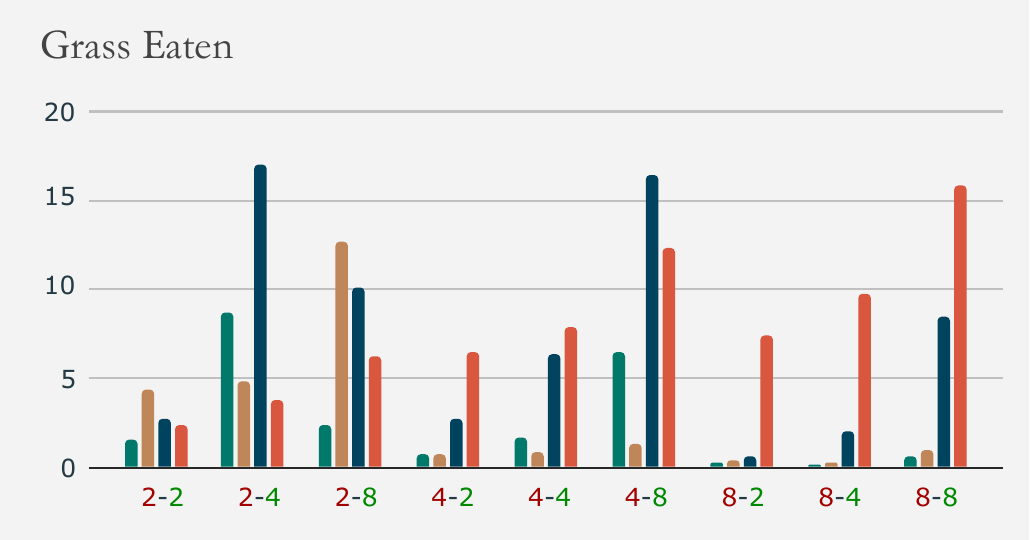}\\
        \includegraphics[width=0.45\linewidth]{ lable.pdf}
        \vspace{-1mm}
        \caption{Grass eating ability of different methods in grassland game. Original Reward and
Reward Engineering sheep both fail to learn to eat grass, especially when there are more wolves. SocialGFs sheep learn to eat
grass successfully, and the amount of grass they eat is inversely proportional to the number of wolves}
        \label{fig:grass}
\end{figure}

In the grassland game, all the reward and grass-eaten statistics are generated by competing against the $SocialGFs$ wolf and sheep in 1000 episodes. 

The sheep have a more challenging role to train in this environment than wolves because 
they must not only avoid wolves but also collect as much grass as possible. In addition to rewards, we also show the grass-eating ability of sheep in the rightmost figure in Fig.~\ref{fig:grass}. This shows how much grass is collected by sheep in 100 time steps.
 We can see that Original Reward and Reward Engineering sheep both fail to learn to eat grass, especially when there are more wolves. This is because being eaten by wolves is a huge punishment for sheep, and the rewards for eating grass are difficult to explore. SocialGFs sheep learn to eat grass successfully, and the amount of grass they eat is inversely proportional to the number of wolves. 

\begin{table}[ht]
\centering
\caption{The occupation rate in cooperative navigation games.}
\resizebox{\textwidth}{!}{
\begin{tabular}{|c|c|c|c|c|c|c|c|c|c|c|c|c|}\hline\hline
   {}&\multicolumn{4}{|c|}{Vanilla Navigation}&\multicolumn{4}{|c|}{Color Navigation}&\multicolumn{4}{|c|}{Team Navigation}\\\hline
   {}&2&3&4&5&2&3&4&5&2&3&4&5\\\hline
      $original reward$& 0.005&0.008&{0.021}&0.031 & 0.02&0.025&{0.028}&0.049 & 0.00&0.002&{0.005}&0.005 \\\hline
      $Reward Engineering$ & 0.189&0.094&{0.082}&0.089 & 0.32&0.13&{0.079}&0.1 & 0.044&0.025&{0.025}&0.026 \\\hline
      $SocialGFs$ & 0.560& \textbf{0.880}& \textbf{0.744}&0.300 & 0.08&0.106&{0.103}&0.101 & 0.0&0.002&{0.003}&0.004 \\\hline
      $SocialGFs^+$ & \textbf{0.570}&{0.876}& \textbf{0.744}&\textbf{0.734} & \textbf{0.719}&\textbf{0.719}& \textbf{0.734}&\textbf{0.719} & \textbf{0.532}&\textbf{0.563}& \textbf{0.583}&\textbf{0.517} \\\hline
      $SocialGFs^*$ & 0.496&0.770&{0.67}&0.668 & 0.651&0.655&{0.685}&0.69 & 0.376&0.440&{0.463}&0.472 \\\hline
\end{tabular}
}

\label{table:occupation}
\end{table}

In cooperation navigation games, we also show occupation rate in table \ref{table:occupation}. It shows how much time a landmark is occupied during the game. We can see from the table that $SocialGF+$ outperforms all the other methods in almost all the tested scenarios. The reward engineering method helps the occupation performance but is not good enough to lead the agents to the success state shown in table\ref{table:navigation}. We run 1000 episodes to calculate the occupation rate and final success rate.

\end{document}